\pdfoutput=1

\documentclass[11pt,table,xcdraw]{article}

\usepackage[]{ACL2023}

\usepackage{times}
\usepackage{latexsym}

\usepackage[T1]{fontenc}

\usepackage[utf8]{inputenc}

\usepackage{microtype}
\usepackage{adjustbox}
\usepackage{graphicx}
\usepackage{booktabs}
\usepackage{multirow}
\usepackage{adjustbox}
\usepackage{caption}
\usepackage{xspace}

\RequirePackage{type1cm}
\RequirePackage{color}
\RequirePackage{soul}
\setstcolor{blue}
\definecolor{violet}{rgb}{0.5,0.0,0.5}
\newsavebox\bscombox
\newcommand{\bscom}[3][]{%
	\sbox{\bscombox}{\fontsize{8}{9}\selectfont#1#2#3}
	\noindent
	\st{#2}{\selectfont
		\color{blue}#3\ifx\\#1\\\else{\fontsize{8}{9}\selectfont\color{violet}[#1]}\fi
	}
}

\newcommand{\distillroberta}{\texttt{paraphrase-distilroberta}\xspace}
\newcommand{\modernbert}{\texttt{ModernBert-base}\xspace}

\usepackage{inconsolata}

\title{Layered Insights: Generalizable Analysis of Human Authorial Style by Leveraging All Transformer Layers}

\author{
	\textbf{Milad Alshomary \textsuperscript{\dag}\textsuperscript{*}},
	\textbf{Nikhil Reddy Varimalla\textsuperscript{\dag}\textsuperscript{\S}\textsuperscript{*}},
	\textbf{Vishal Anand\textsuperscript{\ddag}},
	\\
        \textbf{Smaranda Muresan\textsuperscript{\dag}},
	\textbf{Kathleen McKeown\textsuperscript{\dag}}
	\\
	\\
	\textsuperscript{\dag}Columbia University, New York, USA
	\\
	\textsuperscript{\ddag}Microsoft, Washington, USA
	\\
	\small{
		\textbf{Correspondence:} \href{ma4608@columbia.edu}{ma4608@columbia.edu}
	}
}

\begin{document}
\maketitle

\def\thefootnote{*}\footnotetext{These authors contributed equally to this work}
\def\thefootnote{\S}\footnotetext{Author is currently working at JPMorgan Chase \& Co.}

\def\thefootnote{\arabic{footnote}}

\begin{abstract}
We propose a new approach for the authorship attribution task that leverages the various linguistic representations learned at different layers of pre-trained transformer-based models. We evaluate our approach on two popular authorship attribution models and three evaluation datasets, in in-domain and out-of-domain scenarios. We found that utilizing various transformer layers improves the robustness of authorship attribution models when tested on out-of-domain data, resulting in a much stronger performance. Our analysis gives further insights into how our model's different layers get specialized in representing certain linguistic aspects that we believe benefit the model when tested out of the domain.
\end{abstract}

\section{Introduction}
\label{sec:intro}
Identifying the author of a given text is important for various applications, such as forensic investigation and intellectual property. Authorship attribution (AA) is the task of analyzing the writing style of pairs of texts in order to predict whether they are written by the same author. Approaches developed for this task either focus on explicitly modeling linguistic attributes of texts \cite{koppel2004authorship} or rely on pre-trained transformer-based models to learn relevant features to the task from large training corpora \cite{rivera-soto-etal-2021-learning, wegmann-etal-2022-author}. The latter approach achieves state-of-the-art results on the AA task. However, it focuses on modeling texts by only learning from the final output layer of the transformer, ignoring representations learned at other layers. A large body of research has been investigating the inner workings of pre-trained transformer-based models \cite{rogers-etal-2020-primer}, showing that different layers are specialized in modeling different linguistic phenomena of texts, with lower layers mainly modeling lexical features while higher ones model sentence structure \cite{rogers2021primer}.

\begin{figure}
    \centering
    \includegraphics{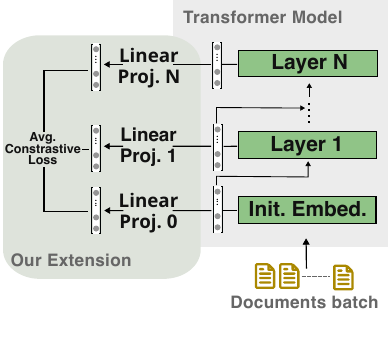}
    \caption{Our approach is a transformer-based model where the initial embeddings (Init. Embed.) and the output of each layer pass through a projection layer, obtaining N+1 document representations that are jointly optimized on the authorship attribution task using contrastive learning}
    \label{fig-our-approach}
\end{figure}

In this paper, we hypothesize that leveraging the representations learned at different layers of a pre-trained transformer can lead to more effective models of AA, especially since the AA task requires modeling similarities between different linguistic aspects of the texts. For example, an author's writing style can be realized by their use of similar vocabulary (lexical), sentence structure (syntax), or even the overall discourse they follow in their writing. To study this hypothesis, we develop a new approach to the task that we call {\bf LIGHT}, providing {\bf l}ayered {\bf i}nsights for the {\bf g}eneralizable analysis of {\bf h}uman authorial style by leveraging all {\bf t}ransformer layers. In particular, as highlighted in Figure \ref{fig-our-approach}, starting from a pre-trained transformer-based model with N layers, our approach projects the initial token embeddings as well as each of the N layers' embedding into a new embedding space, where we apply a contrastive loss that refines this space to better capture the similarity between texts with respect to the original representation. The model is then trained jointly by averaging the contrastive losses at all the N+1 projections. During prediction, given two texts, the analysis of whether they belong to the same author can be decomposed into comparing their similarities at these different learned projections, leveraging more robust predictive power, and allowing a fine-grained analysis of text similarities, a step towards a more interpretable approach to authorship attribution.

To evaluate our hypothesis, we follow the same evaluation setup of \citet{rivera-soto-etal-2021-learning} that considers three datasets for evaluation: Reddit comments, Amazon reviews, and Fanfiction short stories. We apply our approach to two popular AA models: LUAR \cite{rivera-soto-etal-2021-learning} and Wegmann \cite{wegmann-etal-2022-author}, considering two base transformers for each: RoBERTa \cite{liu2019roberta} and ModernBERT \cite{warner2024smarter}. Our experiments demonstrate that, for the out-of-domain scenario, applying our approach on two AA models with two base transformers leads to improvement in 15 out of 16 scenarios, with the highest improvement of around 50\% in the mean reciprocal rank (MRR) for LUAR when trained on the Fanfiction dataset and evaluated on Reddit. Results, however, in the in-domain scenario are mixed. These insights indicate strong generalization capabilities gained when applying our approach, which we argue is more relevant in practical scenarios of the AA task than in-domain. This also provides evidence of the importance of leveraging the various layers of the base transformer when working on the AA task. Furthermore, we perform experiments to analyze the role each layer plays in the overall performance of our model. We found that certain layers became more relevant when models are tested in the out-of-domain setting, leading to high performance. In summary, our contributions are:\footnote{\url{https://github.com/MiladAlshomary/LIGHT-for-AA-analysis}}

\begin{itemize}
    \item A state-of-the-art approach for authorship attribution in out-of-domain scenarios.
    \item Empirical evidence supporting the generalizability of our approach compared to baselines
    \item Insights into the different representations learned in our model
\end{itemize}
\section{Related Work}
\label{sec:related-work}
Transformer-based models have shown strong advances in learning numerous natural language processing tasks. In the pre-training phase, these models are trained on large corpora to predict the next token given a context window, leading them to acquire strong text representations that can be leveraged to learn downstream NLP tasks. In a comprehensive review, \citet{rogers2021primer} showcases how the internal representations of transformers become increasingly specialized across layers, with lower layers capturing surface-level syntactic information and higher layers encoding deeper semantic and task-specific knowledge. \citet{van_Aken_2019} highlights how specific layers in transformers contribute differently to question-answering tasks, underscoring the model's specialization at varying depths. Additionally, \citet{fan2024layersllmsnecessaryinference} provided evidence that only a subset of layers is required for specific downstream tasks, emphasizing the potential for task-dependent optimization of computational resources. These studies collectively emphasize that linguistic knowledge is distributed across layers in a hierarchical and task-dependent manner. Inspired by this literature, in this paper, we study the task of authorship attribution at these different levels of granularity.

\textbf{Author Attribution.} Traditional methods rely on manually designed features, such as syntactic patterns and function word frequencies, to capture distinctive writing styles \cite{koppel2004authorship}. Several datasets were constructed to study the task \cite{potthast2016wrote}. With the rise of deep learning, transformer-based models have set new records by leveraging their ability to capture nuanced stylistic patterns in text. For instance, \citet{rivera-soto-etal-2021-learning} employed a contrastive learning approach to map texts written by the same author into similar regions of an embedding space. 
Similarly, \citet{wegmann-etal-2022-author} introduced methods to ensure that embeddings focus on style rather than content, allowing for more reliable attribution. Despite their success, these approaches rely only on the final layer of the transformer model as the source of text representation. In contrast, our approach leverages different text representations from all the transformer's layers. By applying our approach to these two aforementioned AA models, we demonstrate the gain from modeling the task at different linguistic granularities.

\section{Approach}
\label{sec:approach}
We propose a multi-layer contrastive learning approach that explicitly models authorship attribution at varying levels of linguistic granularity. Instead of relying on a single representation derived from the final layer, our approach extracts embeddings from all the N layers of the transformer, leveraging its linguistic capacity for authorship attribution.

As highlighted in Figure \ref{fig-our-approach}, our approach is transformer-based \cite{vaswani2017attention} with additional N+1 Feed Forward layers that project each of the transformer's hidden states into a new embedding space, including the initial token embeddings. We then apply a contrastive loss function on each of these N+1 embedding spaces to pull documents written by the same author together while keeping other documents further apart in the corresponding space. At each training step, the average loss of all the N contrastive losses is computed and back-propagated throughout the model.
We hypothesize that this architecture ensures that stylistic features across multiple linguistic levels contribute to the final representation, allowing the model to learn complementary signals from different layers. Rather than collapsing stylistic information into a single embedding, we maintain separate representations for each layer. In Section \ref{sec:ling-comp-experiments}, we present an analysis confirming how the various learned layers of our trained model capture different linguistic representations from texts. During inference, given two texts, we compute their similarity at each learned layer projection. These layer-wise similarity scores are aggregated to produce a final similarity measure that determines the likelihood of these two texts being written by the same author. 
\section{Experiment setup}
\begin{table*}[t]
\centering
\begin{tabular}{llrrrrrr}
\multicolumn{2}{c}{{}} & \multicolumn{6}{c}{\textbf{Training Dataset}}\\
\cmidrule{3-8}
\multirow{2}{*}{\textbf{Evaluation Dataset}} & \multirow{2}{*}{\textbf{Model}} & \multicolumn{2}{c}{\textbf{Reddit}} & \multicolumn{2}{c}{\textbf{Amazon}} & \multicolumn{2}{c}{\textbf{Fanfic}} \\
\cmidrule(lr){3-4} \cmidrule(lr){5-6} \cmidrule(lr){7-8}
  & & \textbf{R@8} & \textbf{MRR} & \textbf{R@8} & \textbf{MRR} & \textbf{R@8} & \textbf{MRR} \\
\midrule
\multirow{6}{*}{
    \textbf{Reddit}} 
        & LIGHT-LUAR & \cellcolor{gray!20}67.87 & \cellcolor{gray!20}53.31 & \underline{\textbf{26.92}} & \underline{\textbf{18.03}} & \underline{\textbf{21.11}} & \underline{\textbf{14.32}} \\
        & LUAR & \cellcolor{gray!20}\underline{\textbf{69.67}} & \cellcolor{gray!20}\underline{\textbf{54.78}} & 22.67 & 14.65 & 14.61 & 9.49 \\
        \cmidrule{2-8}
        & LIGHT-Wegmann & \cellcolor{gray!20}\textbf{14.32} & \cellcolor{gray!20}\textbf{9.51} & - & - & - & - \\
        & Wegmann & \cellcolor{gray!20}7.26 & \cellcolor{gray!20}4.83 & - & - & - & - \\
        \cmidrule{2-8}
        & Conv & \cellcolor{gray!20}56.32 & \cellcolor{gray!20}42.38 & 6.30 & 9.70 & 5.74 & 3.90 \\
        & Tf-Idf & \cellcolor{gray!20}10.34 & \cellcolor{gray!20}6.77 & 7.65 & 5.03 & 6.97 & 4.63 \\
\midrule
\multirow{6}{*}{
    \textbf{Amazon}} 
        & LIGHT-LUAR & \textbf{72.63} & \textbf{60.74} & \cellcolor{gray!20}\underline{\textbf{84.68}} & \cellcolor{gray!20}\underline{\textbf{72.65}} & \underline{\textbf{56.69}} & \underline{\textbf{44.36}} \\
        & LUAR & 71.84 & 58.51 & \cellcolor{gray!20}82.54 & \cellcolor{gray!20}68.99 & 37.12 & 26.39 \\
        \cmidrule{2-8}
        & LIGHT-Wegmann & 75.77  & 70.18 & \cellcolor{gray!20}- & \cellcolor{gray!20}- & - & - \\
        & Wegmann & \underline{\textbf{79.49}} & \underline{\textbf{72.26}} & \cellcolor{gray!20}- & \cellcolor{gray!20}- & - & -\\
        \cmidrule{2-8}
        & Conv & 60.20 & 47.60 & \cellcolor{gray!20}74.30 & \cellcolor{gray!20}60.06 & 34.90 & 25.90 \\
        & Tf-Idf & 43.70 & 35.50 & \cellcolor{gray!20}31.61 & \cellcolor{gray!20}24.86 & 21.46 & 16.45 \\
\midrule
\multirow{6}{*}{
    \textbf{Fanfic}} 
        & LIGHT-LUAR & \underline{\textbf{43.26}} & \underline{\textbf{33.96}} & \underline{\textbf{39.21}} & \underline{\textbf{29.98}} & \cellcolor{gray!20}51.78 & \cellcolor{gray!20}44.33 \\
        & LUAR & 42.50 & 32.30 & 34.56 & 25.04 & \cellcolor{gray!20}\underline{{55.38}} & \cellcolor{gray!20}\underline{{45.07}} \\
        \cmidrule{2-8}
        & LIGHT-Wegmann & \textbf{30.65} & \textbf{23.45} & - & - & \cellcolor{gray!20}- & \cellcolor{gray!20}- \\
        & Wegmann      & 28.46 & 21.21 & - & - & \cellcolor{gray!20}- & \cellcolor{gray!20}- \\
        \cmidrule{2-8}
        & Conv & 40.66 & 30.98 & 24.99 & 17.98 & \cellcolor{gray!20}47.98 & \cellcolor{gray!20}39.02 \\
        & Tf-Idf & 25.22 & 18.72 & 26.04 & 19.37 & \cellcolor{gray!20}31.37 & \cellcolor{gray!20}22.53 \\
\bottomrule
\end{tabular}
\caption{Evaluation results comparing the effect of our approach (prefixed with LIGHT) when applied to two AA models, LUAR and Wegmann. Results are shown when all models are evaluated on the three datasets for in-domain (highlighted grey) and out-of-domain settings. The highest value for each model in each setting is highlighted in bold. The overall highest value among all models for each setting is highlighted with underlines.}
\label{single-domain}
\end{table*}

In our evaluation, we closely follow the experiment setup of \citet{rivera-soto-etal-2021-learning} in terms of task definition, datasets, and metrics used. To obtain reliable insights, we test our approach by applying it to two popular AA models: LUAR \cite{rivera-soto-etal-2021-learning} and Wegmann \cite{wegmann-etal-2022-author}, considering two base transformers, RoBERTa \cite{liu2019roberta} and ModernBERT \cite{warner2024smarter}.

\paragraph{Task.} For the authorship attribution task, the input is two collections of texts, each written by a single author, and the output is a score reflecting the likelihood of these two collections being written by the same author. 

\paragraph{Datasets.}
We evaluate the effectiveness of our model on the \textbf{Reddit, Amazon} and \textbf{Fanfiction} datasets used in \citet{rivera-soto-etal-2021-learning}. These datasets represent distinct domains with different linguistic styles, levels of formality, and authorial consistency. Each dataset is partitioned into training and evaluation, ensuring that evaluation texts are temporally disjoint from training data where timestamps are available \cite{andrews-bishop-2019-learning}. For Amazon, we use 135K authors with at least 100 reviews, training on 100K, and evaluating on 35K \cite{ni-etal-2019-justifying}. The Fanfiction dataset \cite{10.1007/978-3-030-58219-7_25} consists of 278,169 stories from 41K authors, with evaluation restricted to 16,456 authors, each contributing exactly two stories \cite{stein:2020k}. The Reddit dataset includes 1 million authors, with training data from \citet{khan-etal-2021-deep} and evaluation using a disjoint, temporally future set \cite{baumgartner2020pushshiftredditdataset}. Further information can be found in Appendix \ref{appendix-data}.

\paragraph{Metrics.}
We follow previous work in evaluating all models using Recall-at-8 (R@8) and Mean Reciprocal Rank (MRR). Recall-at-8 measures the probability that the correct author appears among the top 8 ranked candidates, while MRR evaluates ranking quality based on the position of the first correct author match. Higher values indicate stronger performance \cite{voorhees-tice-2000-trec}.

\paragraph{AA Models.}
As mentioned, we apply our approach by extending two popular AA models, LUAR and Wegmann. \textbf{LUAR} \cite{rivera-soto-etal-2021-learning} is a contrastive learning-based authorship attribution model with a RoBERTa model (\distillroberta checkpoint\footnote{\url{https://huggingface.co/distilbert/distilroberta-base}}) as its base transformer. 
Given the two text collections, the model samples excerpts of 32 tokens and passes them through a RoBERTa transformer, obtaining only the final layer representation of each excerpt. An attention mechanism and max pooling layer are applied to aggregate the extracted representations into a single embedding representing each text collection. A contrastive loss is then applied to these representations to make them closer for positive pairs (written by the same author) and further apart for negative pairs. The \textbf{Wegmann} \cite{wegmann-etal-2022-author} model uses a sentence transformer architecture trained with triplet loss over contrastive anchor-positive-negative sentence triplets to capture authorial style. The core idea of Wegmann is to construct challenging triplets where the negative sentence is from a different author but topically very similar to the anchor author to force the model to learn stylistic differences rather than topical ones. The model is trained on the Reddit dataset, making use of the Subreddit metadata as topical information. We modify each model by applying our approach to extend its architecture with N+1 projection layers. We also consider replacing the RoBERTa transformer base with ModernBERT, a more recent transformer architecture that allows a longer context window (up to 8k tokens) and is trained on a bigger dataset, leading to better performance on various tasks. This also allows us to make sure that our results apply to different transformer architectures. Additionally, we evaluate against the same baselines presented in \citet{rivera-soto-etal-2021-learning}. The first is a convolutional model \cite{andrews-bishop-2019-learning}, which encodes text representations using subword convolutional networks rather than transformers. The second baseline is a TF-IDF baseline, which represents documents as bag-of-words vectors weighted by term frequency in the text and normalized by inverse document frequency.

\begin{table*}[t]
\centering
\begin{adjustbox}{max width=\textwidth}
\begin{tabular}{llcccccc}
\multicolumn{2}{c}{\textbf{}} & \multicolumn{6}{c}{\textbf{Training Dataset}}\\
\cmidrule{3-8}
\multirow{2}{*}{\textbf{Evaluation Dataset}} & \multirow{2}{*}{\textbf{Model}} & \multicolumn{2}{c}{\textbf{Reddit} $\cup$ \textbf{Fanfic}} & \multicolumn{2}{c}{\textbf{Reddit} $\cup$ \textbf{Amazon}} & \multicolumn{2}{c}{\textbf{Amazon} $\cup$ \textbf{Fanfic}} \\
\cmidrule(lr){3-4} \cmidrule(lr){5-6} \cmidrule(lr){7-8}
  & & \textbf{R@8} & \textbf{MRR} & \textbf{R@8} & \textbf{MRR} & \textbf{R@8} & \textbf{MRR} \\
\midrule
\multirow{2}{*}{
    \textbf{Reddit}} 
        & LIGHT-LUAR & \textbf{64.13} & \textbf{49.46} & \textbf{63.96} & \textbf{49.44} & \textbf{29.95} & \textbf{20.62 }\\
        & LUAR & 56.35 & 41.20 & 60.58 & 45.46 & 20.40 & 13.11 \\
\midrule
\multirow{2}{*}{
    \textbf{Amazon}} 
        & LIGHT-LUAR & \textbf{69.43} & \textbf{57.04} & \textbf{89.00} & \textbf{78.38} & \textbf{85.99} & \textbf{74.23} \\
        & LUAR & 53.51 & 39.63 & 84.84 & 72.09 & 79.60 & 64.83 \\
\midrule
\multirow{2}{*}{
    \textbf{Fanfic}} 
        & LIGHT-LUAR & 54.16 & 44.56 & \textbf{43.24} & \textbf{33.54} & \textbf{51.87} & \textbf{42.63} \\
        & LUAR & \textbf{57.51} & \textbf{46.32} & 40.69 & 30.49 & 51.84 & 41.79 \\
\bottomrule
\end{tabular}
\end{adjustbox}
\caption{Evaluation Results after training on multiple domain datasets shown only for LIGHT-LUAR and LUAR}
\label{multi-domain}
\end{table*}

\paragraph{Training.} For \textbf{LUAR}, we follow the training process of \citet{rivera-soto-etal-2021-learning} to train a reproduced version of their LUAR model with RoBERTa base. We train a model for each of the three datasets for 20 epochs, with validation performed every five epochs. We use a batch size of 84 and sample from each author 16 text excerpts of 32 tokens each. For the ModernBERT-based version, we extend training to 30 epochs and 32 text excerpts to accommodate the model's larger size and longer context capacity, respectively, while keeping the excerpt length fixed at 32 tokens and maintaining validation every five epochs. As for \textbf{Wegmann}, we fine-tune the sentence transformer model using triplet loss over 58k triplets constructed from the Reddit training dataset following Wegmann's sampling procedure. Since such topical information is not provided in Amazon or Fanfiction datasets, we refrain from training on them. We train separate models with either RoBERTa or ModernBERT as base transformer, using a batch size of 24 for 20 epochs. Our proposed approach (prefixed as {\bf LIGHT}) extends this by retrieving hidden states from all transformer layers and computing triplet-loss at each layer individually, averaging the resulting losses to promote consistent style alignment throughout the network.

\section{Results}
We examine the impact of our approach when training on single and multiple domains. In the following, we focus on presenting our results considering RoBERTa as the base transformer. Results when using ModernBERT are presented in Appendix \ref{appendix-results} due to space limitations. Note that the findings presented here also apply to ModernBERT.

\begin{figure}[t]
    \centering
    \includegraphics[]{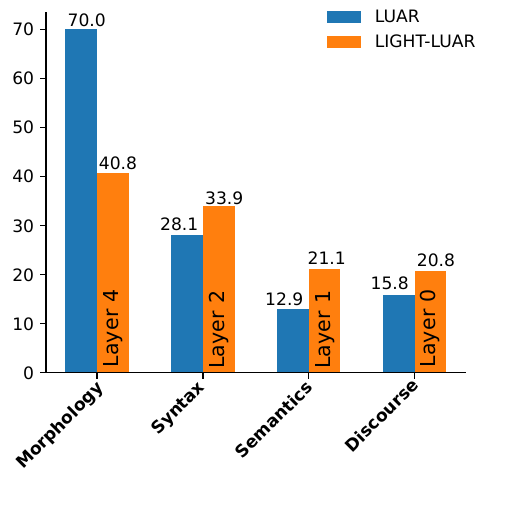}
        \caption{The linguistic capability of LUAR model compared to the best performing layers of our LIGHT-LUAR model trained on the Reddit dataset. }
    \label{fig:ling-capacity-barchart}
\end{figure}

\begin{figure*}[t]
    \centering
    \includegraphics[scale=0.95]{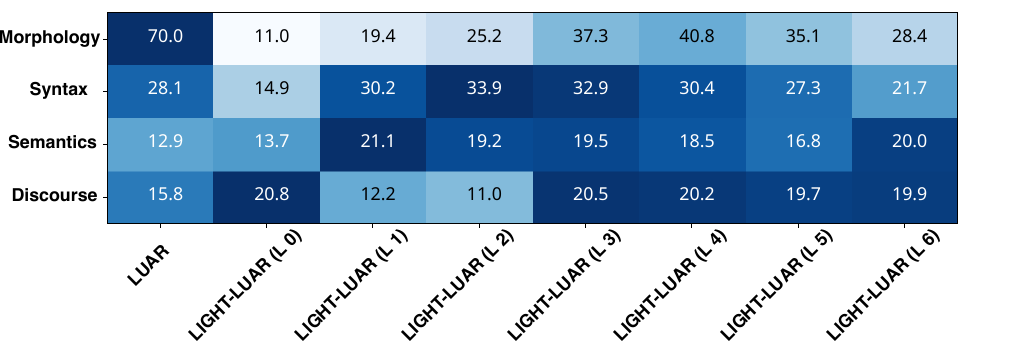}
        \caption{The linguistic capability of each of the 6 projected layers of the LIGHT-LUAR model trained on the Reddit dataset compared to the output layer of the LUAR model (first column) as probed using Holmes benchmark. }
    \label{fig:ling-capacity-roberta}
\end{figure*}

\subsection{Single-Domain Training}
In the single-domain setting, models are trained on a single dataset and tested on the test split of all three datasets to assess their generalizability. Table \ref{single-domain} shows the recall at 8 (R@8) and mean reciprocal rank (MRR) scores. Since training Wegmann's model requires topical information that only exists in the Reddit dataset, the model and our extended version of it (LIGHT-Wegmann) are only trained on Reddit and evaluated on all three datasets.

In general, LUAR and our LIGHT-LUAR model outperform all other baselines in all scenarios except when models are trained on Reddit and evaluated on Amazon. For \textbf{in-domain evaluation}, we observe that applying our approach to LUAR and Wegmann leads to better results only on the Amazon dataset for LUAR and Reddit for Wegmann. In all other in-domain scenarios, applying our approach leads to slightly worse performance. This suggests that for in-domain evaluation, final-layer representations contain sufficient information for authorship attribution, and incorporating multiple layers might lead to a slight degradation in the performance. We argue that the improvement on the Amazon dataset is due to the structured writing style, allowing the different layers to capture more nuanced linguistic representations from texts. In contrast, informal and discussion-based datasets, such as Reddit, exhibit less improvement, likely due to the topic-driven nature of the writing, which can overshadow stylistic signals. This is not the case for LIGHT-Wegmann, which outperforms its baseline (Wegmann) since the model is trained to better separate content from style. 

For \textbf{out-of-domain evaluation}, applying our approach into both AA models demonstrates substantial improvements over baselines, significantly enhancing cross-domain generalization and establishing new state-of-the-art results. LIGHT-LUAR trained on homogeneous datasets, such as Amazon, show remarkable adaptability to free-form writing (An MRR of 18.03 for LIGHT-LUAR compared to 14.65 for LUAR on Reddit dataset), while those trained on Reddit generalize exceptionally well to both Amazon and Fanfiction. However, applying our approach to the Wegmann model leads to improvement in performance in the case of Fanfic but not Amazon. In case of LIGHT-Wegmann with ModernBERT (Table \ref{modernbert-combined-table} in the appendix), our approach is better in all out-of-domain cases. Compared to original AA models, our approach substantially improves transferability, highlighting the importance of leveraging different linguistic granularity levels rather than relying solely on final-layer representations. Most notably, in the Fanfiction transfer setting, where LUAR struggles, our approach achieves 50\% R@8 improvements. 

\subsection{Multi-domain Training}

In this experiment, we train LUAR and our LIGHT-LUAR on two domains and evaluate them on the third one to examine whether exposure to multiple domains enhances robustness. To this end, we train models on domain pairs using mini-batches of 256 that are randomly selected from document collections, ensuring equal representation from both domains. Since authors are disjoint across domains, shared authorship occurs only within the same domain. Table \ref{multi-domain} presents the results. Overall, our LIGHT-LUAR model benefits more than LUAR from including different training domains, achieving higher scores in almost all cases. Compared with the single-domain results shown in Table \ref{single-domain}, we find that introducing Reddit data consistently improves transferability, reinforcing its diverse linguistic coverage as an essential factor in cross-domain generalization. Unlike LUAR, where adding Fanfiction dampened the performance, our LIGHT-LUAR model enhances out-of-domain generalization because it leverages different linguistic granularities. This is evident in models trained on Amazon $\cup$ Fanfiction;
where LUAR failed, LIGHT-LUAR has strong performance on Reddit by integrating multiple stylistic cues. 

\section{Analysis}

In this section, we look at what our model's projection layers capture linguistically and what role they play in the final prediction in both in and out-of-domain scenarios. All analyses are performed on our approach when applied to LUAR.

\subsection{Learned Linguistic Representations}
\label{sec:ling-comp-experiments}

By looking at what linguistic capacities are acquired at the different layers, the goal is to demonstrate the rich linguistic representations captured in our model compared to previous state-of-the-art models. In particular, we use the Holmes benchmark \cite{waldis-etal-2024-holmes}, which consists of over 200 datasets and different probing methods to evaluate the linguistic competence of representations, including syntax, morphology, semantics, and discourse. We run the probing method of this library over each output layer of our LIGHT-LUAR model and on the single output layer of LUAR (both are trained on the Reddit dataset).

Figure \ref{fig:ling-capacity-barchart} shows the best performance of LUAR in representing each of the linguistic phenomena compared to the best performing layers of LIGHT-LUAR. We can observe that LUAR's output layer has a huge skew towards morphology compared to other linguistic phenomena, indicating the importance of morphological representation in texts for the AA task. In contrast, our LIGHT-LUAR model has a more balanced representation of all the linguistic phenomena. More details are shown in Figure \ref{fig:ling-capacity-roberta} as a heatmap of the linguistic competence of all layers of the two models. Aside from Morphology, our LIGHT-LUAR represents each linguistic phenomenon at one of its layers better than LUAR's final output layer. Aligned with previous literature, Syntax is best captured at the intermediate layers (L 2-3), while discourse is at the later layers (L 3-6), as well as L 0, potentially, capturing discourse markers. These observations indicate the need to model authorship attribution at different layers to benefit from the diverse captured linguistic representation, rather than considering only the final output layer.

\begin{figure*}[t]
    \centering
    \includegraphics[width=0.8\textwidth]{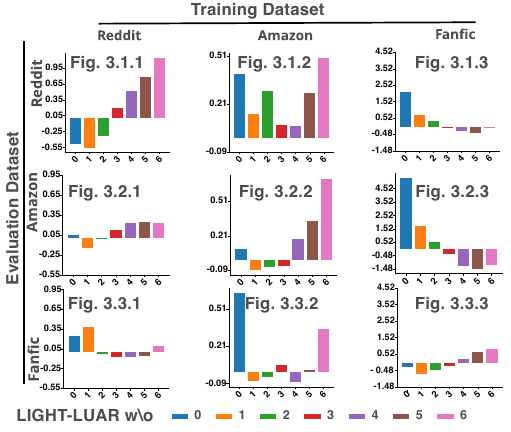}
    \caption{The change in R@8 score in LIGHT-LUAR trained on Reddit, Amazon, and Fanfic when evaluated on all three datasets subject to removing each of its layers. Positive values reflect a drop in performance, while negative values reflect cases where removing the respective layer leads to performance gain. MulitLUAR w\textbackslash o [x] represents the model without the x layer.}
    \label{fig:ablation-study}
\end{figure*}

\subsection{Layer to Prediction Contribution}
We conduct experiments to analyze our model's behavior through an ablation study and by quantifying each layer's significance to the final prediction.

\subsubsection{Ablation Study}
\label{sec:ablation}
The final prediction of our model is an aggregation score of the cosine similarity between all layer representations of the input pairs. Therefore, our ablation study evaluates the impact of different layers in the LIGHT-LUAR model by systematically removing each layer from the aggregation and measuring its effect on attribution performance across three datasets: Reddit, Amazon, and Fanfic, allowing us to categorize this contribution for in and out-of-domain scenarios. Figure \ref{fig:ablation-study} presents the ablation results as a bar chart for each LIGHT-LUAR model when evaluated on each of the three datasets while removing each of its layers from the predictions. Each bar reflects the change in the model's performance in terms of R@8 after removing a specific layer. Negative bars indicate that the layer harmed the model's performance.

\paragraph{LIGHT-LUAR trained on Fanfic} Looking at the final column of Figure \ref{fig:ablation-study}, we notice that when initial layers (Layers 0-3) are removed, there is a significant performance drop in the out-of-domain scenario (Figure 3.2.3 and 3.1.3). This suggests that the lower layers capture stylistic features that generalize well across different datasets. However, this comes with a cost, these initial layers contribute negatively when the model is evaluated in an in-domain setting (Figure 3.3.3), and removing them leads to an increase in the performance. In contrast, the latter layers (4, 5, and 6) contribute more to the in-domain setting; removing them harms the model's performance (Figure 3.3.3).

\paragraph{LIGHT-LUAR trained on Amazon} In the second column of Figure \ref{fig:ablation-study}, we observe a relatively similar behavior. The final layers (4 to 6) contribute mostly to the in-domain setting (Figure 3.2.2). Layers 1 to 3, while benefiting the performance on Reddit (out-of-domain), contribute negatively to in-domain -- removing them improves the performance on the Amazon dataset (Figure 3.2.2).

\paragraph{LIGHT-LUAR trained on Reddit} Now, we consider the first column of Figure \ref{fig:ablation-study}. We observe that removing layer 0 leads to performance degradation for Amazon and Fanfic (out-of-domain) while contributing negatively to the in-domain scenario. Layer 6 then has the highest contribution for the in-domain scenario (Figure 3.1.1).

\begin{figure*}[t]
    \centering
    \includegraphics[]{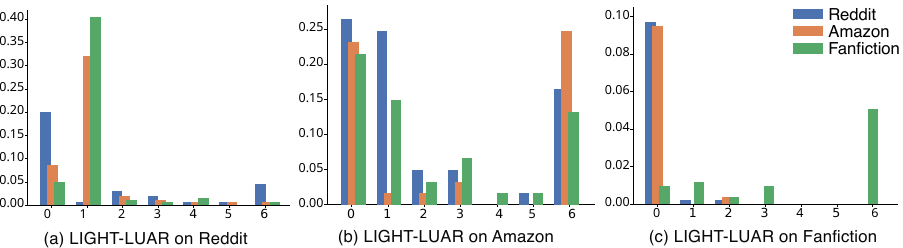}
    \caption{Percentage of text pairs (y-axis) having significantly high similarity score w.r.t. different layers (x-axis) of LIGHT-LUAR trained on (a) Reddit, (b) Amazon, and (c) Fanfiction. These percentages are broken by the evaluation dataset: Reddit, Amazon, and Fanfiction}
    \label{fig:layer-sig-pairs}
\end{figure*}

Overall, the findings reinforce the importance of multi-layer representations in author attribution tasks. Unlike prior work that relies solely on final-layer embeddings, our results demonstrate that early and middle layers' embeddings contribute significantly to out-of-domain generalization, with various layers playing different importance roles based on the evaluation dataset's style characteristics. Higher layer embeddings then capture more nuanced, domain-specific stylistic cues. The LIGHT-LUAR approach, which aggregates information across all linguistic granularity levels, provides a more robust framework for modeling author writing style. These results highlight the limitations of single-layer methods and emphasize the necessity of incorporating multiple levels of linguistic representation for improved attribution performance.

\subsubsection{Layer significance}
We analyze how relevant a specific layer is for predictions on each test dataset (Reddit, Amazon, and Fanfiction). We quantify layer relevance by computing the percentage of instances in the test set (a pair of author texts) with a cosine similarity higher or lower than the average score by a given margin. Specifically, given a corresponding dataset, a LIGHT-LUAR model, and one of its layers, we sample a set of 5k text pairs, measure the cosine similarity of their representations at all layers, and compute the percentage of cases where the z-score of the corresponding layer is higher than 1.5 for positive pairs (indicating cosine similarity 1.5 standard deviations higher than average) or lower than -1.5 for negative pairs (indicating lower cosine similarity than average). We visualize these percentages in Figure \ref{fig:layer-sig-pairs} for the three LIGHT-LUAR models trained on Reddit, Amazon, and Fanfiction datasets. Looking at Figure \ref{fig:layer-sig-pairs}, we can see an overall pattern where the final layer plays a more specific role in predicting in-domain cases while early layers of the model contribute to the predictions of the out-of-domain cases, which aligns with our ablation study presented earlier. For example, considering at LIGHT-LUAR trained on Reddit (Figure 4.a), we observe that layers 0 and 6 are mainly relevant to the Reddit instances, while the other layers, such as layer 1, play a more prominent role in the out-of-domain scenario (Amazon and FanFic datasets). Similarly, for LIGHT-LUAR trained on Amazon (Figure 3.b), both layers 0 and 6 are more relevant to the in-domain predictions (Amazon instances), while for out-of-domain (Reddit and Fanfiction), also early layers such as 1 to 3 contribute to the predictions. Finally, as for the LIGHT-LUAR model trained on Fanfiction (Figure 3.c), we see a clearer pattern where layer 6 contributes mostly to the in-domain predictions while early layers (0 to 2) predict out-of-domain cases.

\section{Discussion and Conclusion}
Starting from the observation that an author's writing style presents itself at different granularities, we proposed a new approach that leverages all the representations learned at different layers of the base transformer to model the AA task more effectively. This feature of our approach becomes more important in out-of-domain scenarios since utilizing multiple levels of linguistic granularity enriches the model's representations, allowing it to perform more robustly. We argue that our approach also allows better interpretability of predictions since authors' similarities can be analyzed on different linguistic levels, allowing a more fine-grained understanding of the prediction.

We evaluated our approach by applying it to two popular AA models and evaluated it on three datasets. Results confirmed how our approach leads to increased performance on the task in the out-of-domain scenario while maintaining relatively strong performance in the in-domain settings. We then studied what linguistic representations were captured at the different layers of our model, demonstrating the reason behind its reliable and robust performance. We further analyze the contribution of each layer in our model towards the final prediction through an ablation study. Across the three trained models, we found a trend of having layers contributing more to the out-of-domain performance than in-domain. Later layers of all models were found to be essential to the in-domain performance. Overall, our findings demonstrate the importance of leveraging all transformer layers when addressing the authorship attribution task to achieve robust performance in out-of-domain.

It is worth drawing parallels between our approach and the classical stylometry approaches for the authorship attribution task \cite{stamatatos2009survey}. While the latter explicitly models the linguistic clues in texts, our approach does this implicitly, where each layer might attend to a different linguistic phenomenon. Future work can evaluate our approach's behavior against these models to have better insights into how much they align in terms of performance. 

\section{Limitations}
Generally, our study focused on the English language, and our findings might not be transferable to other languages. Further research can explore this potential. Moreover, our model architecture implies adding extra parameters to the original AA structure that require more training time and computational resources, especially when the number of layers of the base transformer increases. Future experiments could consider whether all layers need to be projected in order to achieve the best performance. Second, although we verified our hypothesis on two different AA models with two different base transformers, we focused the presentation of our analysis only on the LUAR model due to space limitations. Third, we rely on probing methods to analyze the linguistic competence of our projected layers. However, these might not be very reliable tools, and they have their own limitations.

\section*{Ethics Statement}
We acknowledge that models developed for authorship attribution raise ethical concerns. They can be used to reveal the identity of individuals. While this might be useful for scenarios such as detecting online hate crimes, it could also be used to censor freedom of speech.

Besides, state-of-the-art systems for authorship attribution might suffer from generating false positives, for example, scenarios where a post is wrongly attributed to an individual in a criminal case. Such false positives become more frequent in out-of-domain scenarios. We believe that the research in this paper is a step towards building more robust authorship attribution systems that allow fine-grained analysis of what specific linguistic similarities have been detected, granting more system interpretability.

\section*{Acknowledgements}
This research is supported in part by the Office of the Director of National Intelligence (ODNI), Intelligence Advanced Research Projects Activity (IARPA), via the HIATUS Program contract \#2022-22072200005. The views and conclusions contained herein are those of the authors and should not be interpreted as necessarily representing the official policies, either expressed or implied, of ODNI, IARPA, or the U.S. Government. The U.S. Government is authorized to reproduce and distribute reprints for governmental purposes notwithstanding any copyright annotation therein

\bibliographystyle{acl_natbib}
\bibliography{anthology,custom}

\appendix
\section{Experimental Setup}
\label{appendix-experiment-setup}

\subsection{LUAR}
\paragraph{Model.}
For our experiments, we utilize the RoBERTa model (\distillroberta) \cite{sanh2020distilbertdistilledversionbert} with 82.5M parameters and seven additional linear layers as the backbone for both the LIGHT-LUAR approach and the baseline. We additionally experiment with ModernBERT (\modernbert)\cite{warner2024smarter}, a larger architecture comprising 149M parameters and 23 additional linear layers including embedding layer, which enables deeper stylistic representation and broader contextual modeling across layers. 

Our implementation follows LUAR’s \cite{rivera-soto-etal-2021-learning} approach to memory-efficient attention, which replaces the traditional attention mechanism in the transformer architecture. \citet{rabe2021selfattention}  optimized memory usage by chunking queries, keys, and values into smaller segments, thereby reducing memory overhead without sacrificing computational efficiency. Instead of computing the full attention matrix simultaneously, attention is applied in smaller query chunks (size: 32) and key-value chunks (size: 32). The core optimization is achieved through log-space normalization, where the maximum log value is subtracted before exponentiation to prevent overflow. Attention weights and values are then aggregated in chunks, with final normalization performed only at the end. This method reduces the memory footprint while maintaining the standard O(n²) time complexity, achieving a significantly improved O(log n) memory complexity. Additionally, PyTorch's activation checkpointing is employed, allowing selective recomputation of activations during backpropagation, which further optimizes GPU memory usage.

\paragraph{Training.} Training is conducted using an NVIDIA A100 GPU (40GB memory) to handle the computational demands of large-scale transformer-based models. For RoBERTa-based models, training spans three datasets - Reddit, Fanfiction, and Amazon Reviews, with a fixed schedule of 20 epochs per dataset. Specifically, training on the Reddit dataset requires approximately 1 hour per epoch, while Fanfiction and Amazon take 11 minutes and 5 minutes per epoch, respectively. We do not perform any hyperparameter tuning, opting instead to use the default parameters from the LUAR implementation. Although training beyond 20 epochs yielded minor performance gains, we adhere to the original LUAR setup to ensure reproducibility.

For ModernBERT, training is conducted on the same three datasets with a fixed schedule of 30 epochs per dataset. Due to the model’s increased depth and wider context window, training is significantly slower: the Reddit dataset requires approximately 4 hours per epoch, while Fanfiction and Amazon take 1 hour 10 minutes and 12 minutes per epoch, respectively.

Our training pipeline utilizes PyTorch Lightning (2.5.1.post0), PyTorch (2.5.1+cu121), Transformers (4.51.3), and Scikit-learn (1.6.1), which together facilitate efficient model training, evaluation, and data handling. Since ModernBERT requires Transformers version $\geq$ 4.49.0, we recommend installing PyTorch first, followed by a compatible, upgraded version of Transformers.

\subsection{Wegmann}

\paragraph{Model.}
We evaluate two backbone models: RoBERTa (125M parameters) and ModernBERT (149M parameters), both initialized with pretrained HuggingFace weights. Following the Sentence-BERT framework \cite{reimers-gurevych-2019-sentence}, we implement two variants: a single-layer model, using only the final-layer [CLS] token for embedding; and a proposed all-layer model that incorporates structural modifications to support fine-grained supervision.

The all-layer model uses a custom transformer module (\texttt{TransformerAllLayers}) that forces the underlying transformer to output all intermediate hidden states by enabling \texttt{output\_hidden\_states=True}. These hidden states (a tuple of shape (L, B, T, H) for $L$ layers, batch $B$, tokens $T$, and hidden size $H$) are extracted from all hidden layers and passed through the training pipeline for supervision.

To exploit these, we introduce a \texttt{MultiLayerTripletLoss} objective. For each training triplet (anchor, positive, negative), we extract the [CLS] token from every layer and compute layer-wise cosine distances between anchor-positive and anchor-negative pairs. The loss is defined as $\frac{1}{L} \sum_{\ell=1}^L \max \left(0, d_{\cos}(a^\ell, p^\ell) - d_{\cos}(a^\ell, n^\ell) + m \right)$,
where $a^{\ell}, p^{\ell}$, and $n^{\ell}$ are the [CLS] embeddings from layer $\ell$, and $m$ is the margin (set to 0.5).
This design encourages style-sensitive representations to emerge throughout the network depth, rather than only in the final layer.

\paragraph{Training.}
Training is performed on 59k Reddit-MUD triplets and 210k Wegmann conversation triplets. We use a 4×NVIDIA A100 GPU cluster (80GB) for efficient multi-GPU training. Epoch runtimes vary by model variant and dataset (Reddit and Wegmann data respectively):

\begin{itemize}
\item All-layer: 28 min, 80 min
\item Single-layer: 20 min, 52 min
\end{itemize}

We use early stopping with a $10^{-4}$ threshold and patience of 3 epochs. Training typically converges within 9 epochs. Tokenization edge cases are handled robustly, with input sequences truncated to 512 wordpieces using the HuggingFace tokenizer. Our implementation stack includes PyTorch (2.7.0), Transformers (4.51.3), SentenceTransformers (4.1.0), and Scikit-learn (1.6.1).

\paragraph{Evaluation.}
We follow LUAR’s protocol by ranking each author’s utterance embeddings against a candidate set. However, unlike LUAR, which uses classification logits, Wegmann evaluates directly on sentence embeddings from the trained model (either single-layer or all-layer). Embedding similarities are computed via cosine similarity.

To reduce inference time (during evaluation only), we enable \texttt{torch.bfloat16} precision for both models on A100 hardware. We verified numeric stability by comparing evaluation scores with full-precision outputs and found no degradation in accuracy. This optimization yields an effective $\sim$10× speedup.

Evaluation runtimes (ModernBERT, RoBERTa):
\begin{itemize}
\item Reddit: 24 min, 14 min
\item Fanfiction: 6 min, 4 min
\item Amazon: 120 min, 40 min
\end{itemize}

\subsection{Datasets}
\label{appendix-data}
Our study is based on publicly available datasets, which have been widely used in prior research on author attribution and stylistic analysis. These datasets originate from online platforms and may inherently contain offensive or personally identifiable content. However, we do not apply any additional filtering beyond what has been done in previous studies. To ensure ethical usage, we follow the guidelines set by the original sources that published these datasets and acknowledge any potential biases or content-related concerns that may arise.

Additionally, we discuss the licensing terms associated with the datasets used in our experiments. The datasets are provided under open-access licenses, allowing their use for academic research. We ensure compliance with these licenses and properly attribute the sources in our work. Detailed information on dataset licensing is included in the final version of the paper.

\section{ModernBERT Results}
\label{appendix-results}
We evaluated our methodology using ModernBERT as the foundational transformer model. The outcomes demonstrate consistent and significant improvements over the RoBERTa-based models discussed previously, attributable to ModernBERT's increased context length and greater model depth. Results are shown in Table \ref{modernbert-combined-table}

For \textbf{in-domain evaluation}, the performance patterns align closely with previous observations using RoBERTa. Incorporating multiple transformer layers with ModernBERT-based LUAR and Wegmann approaches typically results in comparable or marginally reduced performance relative to their single-layer counterparts. This indicates that within the same domain, the deeper final-layer representations from ModernBERT already encapsulate extensive stylistic and content information, minimizing the benefits derived from incorporating intermediate layers. Nonetheless, we note distinct improvements in scenarios involving structured writing styles, such as the Amazon dataset, where leveraging multiple layers evidently captures subtle stylistic features more effectively than using the final layer alone.

In the \textbf{out-of-domain evaluation} setting, however, our multi-layer approach with ModernBERT substantially enhances model generalizability and sets new benchmarks compared to the previously used RoBERTa-based models. The expanded contextual understanding and richer intermediate representations provided by ModernBERT significantly improve cross-domain adaptability. For instance, ModernBERT-based LIGHT-LUAR trained on structured datasets like Amazon demonstrates notably superior adaptability when evaluated on informal, conversational texts like Reddit and Fanfiction, capturing nuanced stylistic features that generalize across diverse domains. Similarly, when models trained on conversational datasets such as Reddit are evaluated on structured writing domains (Amazon), the ModernBERT variants consistently outperform previous baselines.

Most notably, on the challenging Fanfiction dataset, known for its diverse and expressive language usage, our ModernBERT-based LIGHT-LUAR and LIGHT-Wegmann models exhibit remarkable improvements, highlighting their ability to generalize across significantly different linguistic styles. These improvements underscore the value of leveraging multiple transformer layers in ModernBERT, particularly for enhancing performance in complex, stylistically rich domains.
\begin{table*}[t]
\centering
\begin{tabular}{llrrrrrr}
\multicolumn{2}{c}{} & \multicolumn{6}{c}{\textbf{Training Dataset}}\\
\cmidrule{3-8}
\multirow{2}{*}{\textbf{Evaluation Dataset}} & \multirow{2}{*}{\textbf{Model}} & \multicolumn{2}{c}{\textbf{Reddit}} & \multicolumn{2}{c}{\textbf{Amazon}} & \multicolumn{2}{c}{\textbf{Fanfic}} \\
\cmidrule(lr){3-4} \cmidrule(lr){5-6} \cmidrule(lr){7-8}
  & & \textbf{R@8} & \textbf{MRR} & \textbf{R@8} & \textbf{MRR} & \textbf{R@8} & \textbf{MRR} \\
\midrule
\multirow{6}{*}{\textbf{Reddit}} 
        & LIGHT-LUAR & \cellcolor{gray!20}70.37 & \cellcolor{gray!20}55.49 & \underline{\textbf{26.77}} & \underline{\textbf{17.85}} & \underline{\textbf{16.26}} & \underline{\textbf{10.86}} \\
        & LUAR & \cellcolor{gray!20}\underline{\textbf{72.29}} & \cellcolor{gray!20}\underline{\textbf{56.89}} & 19.88 & 12.70 & 10.95 & 6.81 \\
        \cmidrule{2-8}
        & LIGHT-Wegmann & \cellcolor{gray!20}\textbf{13.91} & \cellcolor{gray!20}\textbf{9.44} & - & - & - & - \\
        & Wegmann & \cellcolor{gray!20}5.90 & \cellcolor{gray!20}3.79 & - & - & - & - \\
        \cmidrule{2-8}
        & Conv & \cellcolor{gray!20}56.32 & \cellcolor{gray!20}42.38 & 6.30 & 9.70 & 5.74 & 3.90 \\
        & Tf-Idf & \cellcolor{gray!20}10.34 & \cellcolor{gray!20}6.77 & 7.65 & 5.03 & 6.97 & 4.63 \\
\midrule
\multirow{6}{*}{\textbf{Amazon}} 
        & LIGHT-LUAR & \underline{\textbf{87.65}} & \underline{\textbf{80.79}} & \cellcolor{gray!20}\underline{\textbf{96.70}} & \cellcolor{gray!20}\underline{\textbf{92.07}} & \underline{\textbf{63.52}} & \underline{\textbf{50.65}} \\
        & LUAR & 86.38 & 78.69 & \cellcolor{gray!20}95.90 & \cellcolor{gray!20}89.12 & 32.39 & 22.14 \\
        \cmidrule{2-8}
        & LIGHT-Wegmann & \textbf{82.34}  & \textbf{77.64} & \cellcolor{gray!20}- & \cellcolor{gray!20}- & - & - \\
        & Wegmann & 79.72 & 71.99 & \cellcolor{gray!20}- & \cellcolor{gray!20}- & - & -\\
        \cmidrule{2-8}
        & Conv & 60.20 & 47.60 & \cellcolor{gray!20}74.30 & \cellcolor{gray!20}60.06 & 34.90 & 25.90 \\
        & Tf-Idf & 43.70 & 35.50 & \cellcolor{gray!20}31.61 & \cellcolor{gray!20}24.86 & 21.46 & 16.45 \\
\midrule
\multirow{6}{*}{\textbf{Fanfic}} 
        & LIGHT-LUAR & \underline{\textbf{46.32}} & \underline{\textbf{35.83}} & \underline{\textbf{39.71}} & \underline{\textbf{29.01}} & \cellcolor{gray!20}56.50 & \cellcolor{gray!20}46.46 \\
        & LUAR & 42.41 & 31.79 & 27.86 & 18.64 & \cellcolor{gray!20}\underline{\textbf{58.47}} & \cellcolor{gray!20}\underline{\textbf{47.67}} \\
        \cmidrule{2-8}
        & LIGHT-Wegmann & \textbf{36.33} & \textbf{28.50} & - & - & \cellcolor{gray!20}- & \cellcolor{gray!20}- \\
        & Wegmann & 24.39 & 18.14 & - & - & \cellcolor{gray!20}- & \cellcolor{gray!20}- \\
        \cmidrule{2-8}
        & Conv & 40.66 & 30.98 & 24.99 & 17.98 & \cellcolor{gray!20}47.98 & \cellcolor{gray!20}39.02 \\
        & Tf-Idf & 25.22 & 18.72 & 26.04 & 19.37 & \cellcolor{gray!20}31.37 & \cellcolor{gray!20}22.53 \\
\bottomrule
\end{tabular}
\caption{Evaluation results using ModernBERT-based LUAR (LU) and Wegmann (WG) variants, compared against Conv and Tf-Idf baselines. In-domain results are highlighted in gray. Highest values within pairs are bolded; global bests are underlined.}
\label{modernbert-combined-table}
\end{table*}

\section{Ablation Study}
\subsection{Significance of Layer 0}

\begin{table*}[t]
\centering
\begin{tabular}{llcccccc}
        \toprule
        \multirow{2}{*}{Evaluation Dataset} & \multirow{2}{*}{Model} & \multicolumn{2}{c}{Reddit} & \multicolumn{2}{c}{Amazon} & \multicolumn{2}{c}{Fanfic} \\
        \cmidrule(lr){3-4} \cmidrule(lr){5-6} \cmidrule(lr){7-8}
        & & R@8 & MRR & R@8 & MRR & R@8 & MRR \\
        \midrule
        \multirow{2}{*}{Reddit} 
        & LIGHT-LUAR w\ [0]  & 67.87 & 53.31 & \textbf{26.92} & \textbf{18.03} & \textbf{21.11} & \textbf{14.32} \\
        & LIGHT-LUAR w\o [0] & \textbf{68.55} & \textbf{53.7} & 26.76 & 17.94 & 19.38 & 12.73 \\
        \midrule
        \multirow{2}{*}{Amazon} 
        & LIGHT-LUAR w\ [0]  & \textbf{72.63} & \textbf{60.74} & \textbf{84.68} & \textbf{72.65} & \textbf{56.69} & \textbf{44.36} \\
        & LIGHT-LUAR w\o [0] & 72.14 & 59.58 & 84.12 & 71.78 & 52.97 & 40.51 \\
        \midrule
        \multirow{2}{*}{Fanfic} 
        & LIGHT-LUAR w\ [0]  & \textbf{43.26} & \textbf{33.96} & \textbf{39.21} & \textbf{29.98} & \textbf{51.78} & \textbf{44.33} \\
        & LIGHT-LUAR w\o [0] & 42.84 & 33.44 & 38.64 & 28.65 & 51.41 & 42.28 \\
        \bottomrule
    \end{tabular}
    \caption{Performance comparison of LIGHT-LUAR with and without [0] across different training datasets. Bold values indicate the better performance between the two variations.}
    \label{tab:layer-0}
\end{table*}

Our model,LIGHT-LUAR, consists of seven projection layers, where \textbf{Layer 0} represents the projection of the initial token embeddings from the \texttt{distillroberta} transformer, while \textbf{Layers 1 to 6} correspond to the projection of its six hidden layers. To demonstrate that, although Layer 0 is not part of the model architecture, it plays a significant role in authorship attribution, we train and evaluate all models on the three datasets without adding a linear projection to Layer 0. The embedding is not incorporated into the loss calculation and similarity metric computation. The results in Table~\ref{tab:layer-0} compare the performance ofLIGHT-LUAR with and without Layer 0 across different evaluation datasets. We observe that including Layer 0 generally improves performance in out-of-domain evaluation, particularly on the Amazon and Fanfiction datasets, where it enhances R@8 and MRR scores. This suggests that initial token embeddings contain valuable stylistic information that contributes to generalization across domains. However, for in-domain evaluation on Reddit, the model without Layer 0 slightly outperforms the full model as shown in section \ref{sec:ablation}, indicating that deeper layers are more influential in capturing dataset-specific stylistic patterns. These findings reinforce the importance of leveraging multi-layer representations for robust authorship attribution.

\end{document}